\documentclass[conference]{IEEEtran}

\IEEEoverridecommandlockouts

\usepackage{times}

\usepackage[numbers]{natbib}
\usepackage{multicol}
\usepackage[bookmarks=true, hidelinks]{hyperref}

\usepackage{graphicx}
\usepackage{mathnotation}
\usepackage{multirow}

\begin{document}

\title{Geometric Gait Optimization for Kinodynamic Systems Using a Lie Group Integrator}

\author{Yanhao Yang, Ross L. Hatton %
\thanks{This work was supported in part by NSF Grants No. 1653220, 1826446, and 1935324, and by ONR Grant No. N00014-23-1-2171.} %
\thanks{Y. Yang and R. L. Hatton are with the  Collaborative Robotics and Intelligent Systems (CoRIS) Institute at Oregon State University, Corvallis, OR USA. {\tt\small \{yangyanh, Ross.Hatton\}@oregonstate.edu}} %
}

\maketitle

\begin{abstract}
This paper presents a gait optimization and motion planning framework for a class of locomoting systems with mixed kinematic and dynamic properties. Using Lagrangian reduction and differential geometry, we derive a general dynamic model that incorporates second-order dynamics and nonholonomic constraints, applicable to kinodynamic systems such as wheeled robots with nonholonomic constraints as well as swimming robots with nonisotropic fluid-added inertia and hydrodynamic drag. Building on Lie group integrators and group symmetries, we develop a variational gait optimization method for kinodynamic systems. By integrating multiple gaits and their transitions, we construct comprehensive motion plans that enable a wide range of motions for these systems. We evaluate our framework on three representative examples: roller racer, snakeboard, and swimmer. Simulation and hardware experiments demonstrate diverse motions, including acceleration, steady-state maintenance, gait transitions, and turning. The results highlight the effectiveness of the proposed method and its potential for generalization to other biological and robotic locomoting systems.
\end{abstract}

\IEEEpeerreviewmaketitle

\section{Introduction}

\begin{figure}[!t]
\centering
\includegraphics[width=\linewidth]{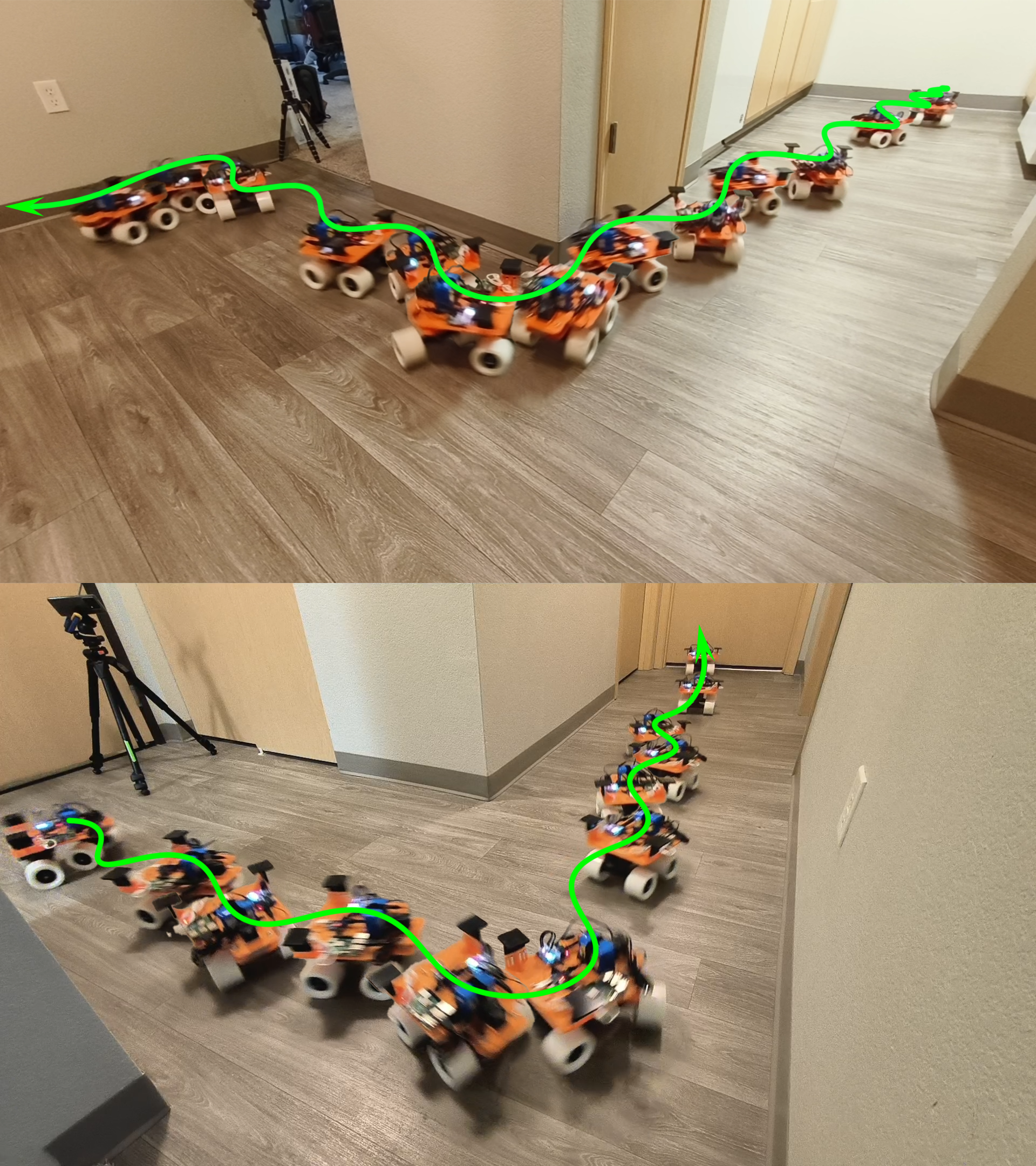}
\caption{Execution of the comprehensive motion plan computed using the proposed method on the roller racer robot. The green curve depicts the robot's position trajectory, and the composite figure illustrates the robot's configuration at different times. The roller racer has only one actuated degree of freedom: the steering angle between the front and rear wheels, with all wheels being passive and driven by anti-slip constraints and steering angle changes. The motion plan incorporates multiple gaits and transitions, enabling the robot to: a) start from rest and nominal shape into the gait cycle, b) accelerate to the optimal steady-state motion, c) smoothly switch to the steady-state gait that maintains maximum speed, d) execute a turn with minimal energy loss and return to steady-state motion, and e) end the gait cycle and return to the nominal shape. Throughout the execution, the motion plan maintains the desired heading angle and meets the average power consumption constraints.}
\label{fig:intro}
\end{figure}

Animals use a diverse set of gaits---such as horses' trotting and galloping, snakes' slithering and sidewinding, and body-caudal fin swimming in fish---to locomote effectively in various environments. By switching or transitioning between different gaits, animals demonstrate remarkable efficiency and adaptability, even in unstructured terrains \cite{alexander2006principles}.

In both biology and robotics, significant research has been conducted on various forms of locomotion. Beyond systems whose dynamics are either fully determined by nonholonomic constraints or solely governed by conservation of momentum \cite{shammas2007geometric}, there exist classes of locomoting systems that exhibit both kinematic and dynamic effects. These systems, referred to in this paper as kinodynamic systems, have garnered considerable attention due to their time-varying nonlinear dynamics and high degree of underactuation.

A representative example of kinodynamic systems is aquatic animals swimming at intermediate Reynolds numbers. The swimming of these aquatic animals is modeled as a system with drift (the ability to glide) due to the mixed inertial and fluid drag effects \cite{mason2003fluid, kelly2012geometric}. This system is distinct from swimming at very high or low Reynolds numbers, where the dynamics are respectively dominated by conservation of momentum \cite{kanso2009swimming, hatton2022geometry} or quasistatic drag equilibrium \cite{mason2003fluid, xiong2007geometric, ramasamy2019geometry}, both of which produce driftless (or ``kinematic'') locomotion behavior. There is existing work that applies perturbation analysis to kinodynamic swimming locomotion to approximate the average motion over a gait cycle, enabling the design of sinusoidal gaits and image-based feedback control \cite{mcisaac2003motion}. Subsequent study models robotic fish by incorporating fluid-added mass, lift, and drag forces, deriving oscillatory control strategies through controllability analysis \cite{morgansen2007geometric}. Other works extend Lighthill’s large-amplitude elongated body theory by introducing resistive forces \cite{boyer2008fast, boyer2010poincare}, resulting in a kinodynamic swimming model, which was later used in \cite{porez2014improved} for modeling swimming robots and performing gait search. Additionally, simplified swimming robot systems, such as Chaplygin’s sleigh and its variants, have been studied for their control and motion planning \cite{kelly2012proportional, fedonyuk2020path, knizhnik2021thrust}.

Interestingly, some wheeled robots are also kinodynamic systems, characterized by their dynamics governed by a combination of nonholonomic constraints and nonholonomic momentum evolution. The snakeboard \cite{ostrowski1995mechanics, ostrowski1998reduced} and roller racer \cite{krishnaprasad2001oscillations, halvani2022nonholonomic} are two representative examples. These systems have inspired extensive research on kinodynamic locomotion, including controllability analysis and gait design using sinusoidal waves \cite{murray1993nonholonomic, ostrowski1994nonholonomic, krishnaprasad2001oscillations}, gait selection via optimal control theory \cite{ostrowski2000optimal}, and motion planning formulated as a nonlinear inversion problem \cite{bullo2001kinematic, bullo2003kinematic}. Other studies have explored gait design using area rules \cite{shammas2007unified}, analytical solutions for trajectory tracking \cite{shammas2012motion, dear2013snakeboard}, and the integration of kinodynamic systems with bipedal riders \cite{anglingdarma2021motion}. Additionally, efforts have been made to enhance the fidelity of kinodynamic models based on the snakeboard and roller racer. For instance, some studies incorporate dissipative forces or viscous friction \cite{kelly1996geometry, ostrowski1998reduced, krishnaprasad2001oscillations}, account for skidding effects \cite{dear2015snakeboard}, or introduce improved friction modeling \cite{pengcheng2006motion, salman2016physical}. Variants of these systems have also been studied, with research focusing on their modeling and control \cite{chitta2005robotrikke, shammas2007unified}.

Given that the positions of kinodynamic systems relative to the world are described by elements of Lie groups, numerical methods that exploit this group structure and preserve its symmetries offer significant advantages in terms of accuracy and efficiency \cite{marsden2001discrete, hairer2006geometric, celledoni2014introduction}. These methods have demonstrated superior performance across various domains, including simulation \cite{jonghoonpark2005geometric, bruls2010use}, control \cite{kobilarov2009lie, wu2015safetycritical}, and estimation \cite{barrau2017invariant, barrau2018invariant}. Moreover, variational integrators and discretization techniques that preserve group structures can be combined with optimal control to enhance both accuracy and stability \cite{leok2007overview, bloch2008optimal, kobilarov2011discrete, jimenez2013discrete, saccon2013optimal}. This idea has been applied to a variety of robotic systems, including ground vehicles \cite{kobilarov2009lie}, quadrotors \cite{kobilarov2014discrete}, and legged robots \cite{teng2022errorstate, csomay-shanklin2023nonlinear}. These advancements highlight the potential of Lie group integrators as a powerful tool for gait optimization in kinodynamic systems.

A common baseline for gait optimization is to use standard optimal control techniques---such as direct collocation \cite{chang2018optimal} or indirect methods \cite{wiezel2023geometric}---without exploiting Lie group theory. However, direct collocation suffers from discretization errors, in which quadrature-based integral approximations may violate dynamic constraints and accumulate errors, allowing the optimizer to ``cheat'' in highly nonlinear systems. While indirect methods offer better forward rollout accuracy, their backward pass is more complex and they struggle to incorporate additional state or input constraints. In contrast, gait optimization methods that leverage Lie group theory enable full-fidelity simulations and allow for the direct imposition of constraints on both control inputs and trajectories, with gradients obtained via the chain rule.

One line of related work uses Lie group theory for geometric gait optimization via constraint curvature and the generalized Stokes' theorem \cite{ramasamy2019geometry, hatton2022geometry}, but these methods mainly apply to principally kinematic systems. Geometric insights into the dynamic phase in kinodynamic systems have been developed to better understand their structure, though the approach relies on trial-and-error selection of gaits that couple the geometric and dynamic phases \cite{shammas2007unified}. Furthermore, other studies employing Lie group theory or integrators for motion planning in kinodynamic systems have predominantly focused on waypoint tracking or trajectory optimization \cite{leok2007overview, bloch2008optimal, kobilarov2011discrete, jimenez2013discrete, saccon2013optimal}. These methods can suffer from discretization errors over long trajectories and require recalculation for each new waypoint or trajectory. In contrast, this work focuses on gait optimization to avoid long trajectory discretization and to generate reusable motion primitives that enable comprehensive motion planning.

The main contribution of this paper is a gait optimization and motion planning framework for kinodynamic systems. This framework comprises:
\begin{itemize}
    \item A general kinodynamic model incorporating kinematic and dynamics effects, based on Lagrangian reduction and differential geometry, applicable to a broad range of kinodynamic systems, including wheeled robots with nonholonomic constraints and swimming robots with anisotropic fluid-added inertia and hydrodynamic drag.
    \item A variational gait optimization method for kinodynamic systems, leveraging Lie group integrators and group symmetries.
    \item A methodology for constructing comprehensive motion plans that integrate multiple gaits and their transitions, enabling diverse motion capabilities for kinodynamic systems.
\end{itemize}
We evaluate our approach on three representative systems: the roller racer, the snakeboard, and a swimmer at an intermediate Reynolds number. All systems are evaluated in simulation, and the roller racer is additionally tested in hardware experiments. These evaluations demonstrate the ability of our framework to achieve a wide range of motions, including acceleration, steady-state maintenance, gait transitions, and turning, as shown in Fig.~\ref{fig:intro} that illustrates the experiment conducted on the roller racer robot. The results underscore the effectiveness of the proposed approach and its potential for generalization to other biological and robotic locomotion systems.

\section{Kinodynamic System Modeling}

\begin{figure*}[!t]
\centering
\includegraphics[width=\linewidth]{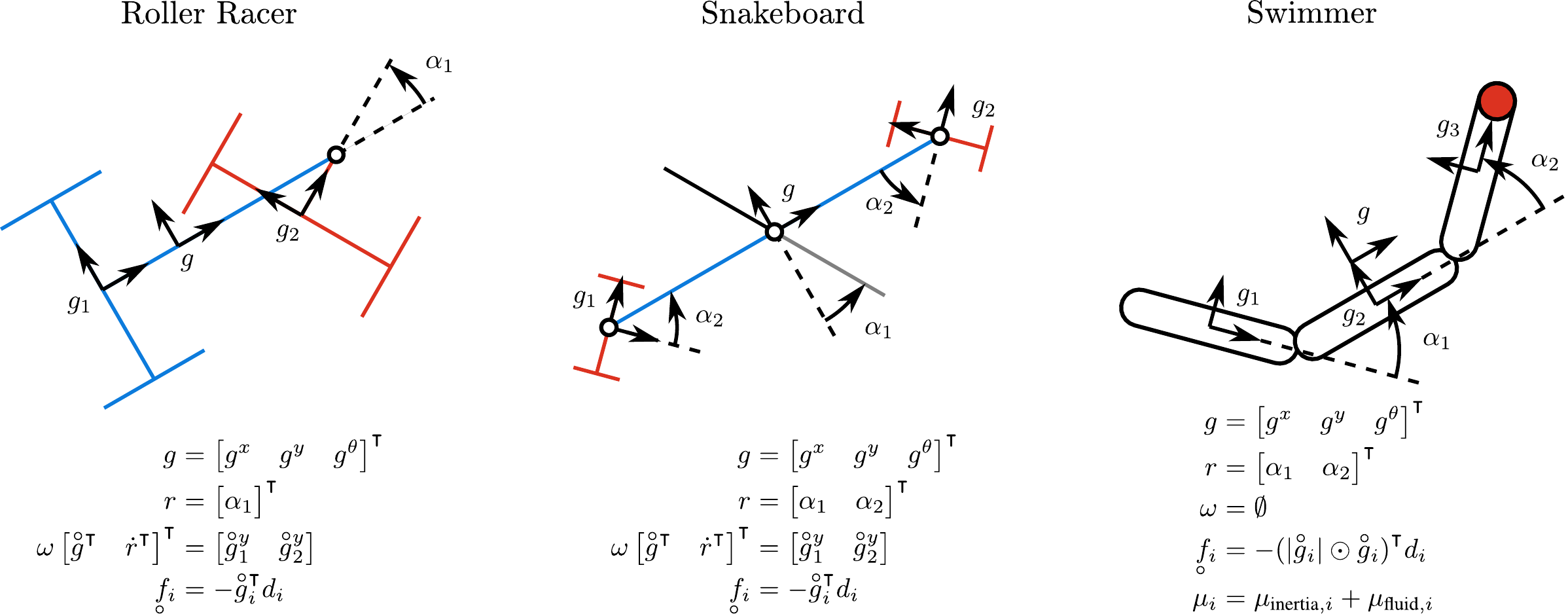}
\caption{Illustration and key characteristics of the three kinodynamic systems considered in this paper. The roller racer has four degrees of freedom (DOF): three DOFs for position, describing its pose in the world, and one DOF for shape (the heading angle between the two carts). The system includes two nonholonomic constraints, corresponding to the non-lateral slip constraints of the wheelsets on the two carts, resulting in one DOF of nonholonomic momentum. Linear viscous friction occurs in the driving direction of each wheelset. Similarly, the snakeboard has five DOFs: three DOFs for position and two DOFs for shape. The shape is defined by the rotor angle and the cart angle, with the assumption that the two carts rotate together in opposite directions. The two nonholonomic constraints, corresponding to the non-lateral slip constraints of the wheelsets, also result in one DOF of nonholonomic momentum. Linear viscous friction also acts in the driving direction of each wheelset. Finally, the intermediate Reynolds number swimmer has five DOFs: three DOFs for position and two DOFs for shape, defined by the joint angles between its three links. Unlike the other systems, it has no direct nonholonomic constraints, so its momentum corresponds to the three DOFs for position. Instead, fluid dynamics manifest as anisotropic fluid-added mass and hydrodynamic drag acting on each link.}
\label{fig:systems}
\end{figure*}

Kinodynamic systems are characterized by the interplay of kinematic and dynamic effects. Kinematic effects include the relationship between inertia, velocity, and momentum, as well as nonholonomic constraints. Based on the kinematics, one can derive the reconstruction equation that relates the system's body velocity to its shape velocity and nonholonomic momentum. Dynamic effects are governed by the Euler-Lagrange equations, incorporating dissipative or frictional forces and constrained forces. Through the application of Lagrangian reduction, these dynamics can be simplified into nonholonomic momentum evolution equations \cite{bloch1996nonholonomic}.

\subsection{Reconstruction Equation}

For a general locomoting system, the Lagrangian, $\lagrangian$, is defined as
\begin{eqalignbracket}
    \lagrangian(\base, \basedot, \fiber, \fibercirc) &= \kineticenergy(\base, \basedot, \fibercirc) - \potentialenergy(\base) \\
    \kineticenergy(\base, \basedot, \fibercirc) &= \sum_{i}\frac{1}{2}\transpose{\fibercirc}_{i}\inertiametric_{i}\fibercirc_{i} \\
    &= \frac{1}{2}\transpose{\begin{bmatrix}
        \fibercirc \\ \basedot
    \end{bmatrix}}\begin{bmatrix}
        \Inertiametric_{\fiber\fiber} & \Inertiametric_{\fiber\base} \\ \Inertiametric_{\base\fiber} & \Inertiametric_{\base\base}
    \end{bmatrix}\begin{bmatrix}
        \fibercirc \\ \basedot
    \end{bmatrix},
\end{eqalignbracket}
where $\kineticenergy$ and $\potentialenergy$ represent the kinetic and potential energy, respectively.\footnote{For simplicity, this paper focuses on changes in kinetic energy during locomotion, assuming constant potential energy. This assumption is based on gravity being counteracted by the ground reaction force in wheeled robots or by neutral buoyancy in swimmers. However, the proposed approach is generalizable to systems with variable potential energy, such as those influenced by gravity or joint stiffness.} The terms $\base$ and $\basedot$ correspond to the system's shape and shape velocity, while $\fiber$ and $\fibercirc$ denote the system's position in $SE(2)$ and its body velocity in $\mathfrak{se}(2)$, respectively.\footnote{The ``open circle'' notation we use here is similar to the ``dot'' notation, but denotes derivatives taken in the body frame. Formally, these are derivatives with respect to right-group actions (locally with respect to the current configuration) rather than coordinate values, producing left-invariant representations of velocity and force. The ``open circle'' superscripts represent tangent vectors, such as velocity, and the subscripts are for covectors such as force or momentum. By convention, velocity is represented as the time derivative of the configuration, and momentum and force are given their own symbols. 

For example, the full reading of the notation of the body velocity $\fibercirc$ is ``the velocity constructed by taking the derivative of position with respect to right group actions'', and the full reading of the notation of body force $\coforce$ is ``the force constructed by taking the derivative of dissipated energy with respect to right (body) velocities''.} $\Inertiametric$ represents the inertia matrix, with the subscripts representing the entries corresponding to the coordinates. $\fibercirc_{i}$ and $\inertiametric_{i}$ indicating the body velocity and inertia matrix of the $i$-th component of the system in its body frame. 

Based on the Lagrangian, we can define the body momentum, $\bodymom$, and the momentum in the generalized coordinates of the system shape, $\basemom$, as\footnote{Momentum is a covector, so we represent it as a row vector here.}
\begin{eqalignbracket}
    \begin{bmatrix}
        \bodymom & \basemom
    \end{bmatrix} &= \begin{bmatrix}
        \parderiv{\lagrangian}{\fibercirc} & \parderiv{\lagrangian}{\basedot}
    \end{bmatrix} \\
    &= \begin{bmatrix}
        \transpose{\fibercirc} & \transpose{\basedot}
    \end{bmatrix}\begin{bmatrix}
        \Inertiametric_{\fiber\fiber} & \Inertiametric_{\fiber\base} \\ \Inertiametric_{\base\fiber} & \Inertiametric_{\base\base}
    \end{bmatrix}.
\end{eqalignbracket}

If any nonholonomic constraints exist, they can be encoded by a linear operator $\nonholonomic$ which maps permitted combinations of body and shape velocity to zero
\beqbracket
    \nonholonomic(\base)\begin{bmatrix}
        \fibercirc \\ \basedot
    \end{bmatrix}_{\text{permitted}} = 0.
\eeqbracket
Then, the total nonholonomic momentum, $\bar{\genmom}$, is defined as the projection of the system's momentum onto the allowable motion space, determined by the kernel of the constraint map~$\bar{\nonhmap}$
\begin{eqalign}
    \bar{\nonhmap} &= \ker(\nonholonomic) \\
    \bar{\genmom} &= \left\langle\begin{bmatrix}
        \bodymom & \basemom
    \end{bmatrix}; \bar{\nonhmap}\right\rangle.
\end{eqalign}
If no nonholonomic constraints exist in the system, we can simply skip this step and define $\nonholonomic = \emptyset$ and $\bar{\nonhmap} = \matrixid$.

By the definition of nonholonomic momentum and nonholonomic constraints, we can derive the system kinematics, which relates the system's body and shape velocities to the current nonholonomic momentum and nonholonomic constraints:
\beqbracket
    \begin{bmatrix}
        \fibercirc \\ \basedot
    \end{bmatrix} = \inv{\begin{bmatrix}
        \transpose{\bar{\nonhmap}}\begin{bmatrix}
            \Inertiametric_{\fiber\fiber} & \Inertiametric_{\fiber\base} \\ \Inertiametric_{\base\fiber} & \Inertiametric_{\base\base}
        \end{bmatrix} \\
        \nonholonomic
    \end{bmatrix}}\begin{bmatrix}
        \transpose{\bar{\genmom}} \\
        0
    \end{bmatrix}.
\eeqbracket
In this paper, we assume direct control of the system shape, meaning that unit vectors corresponding to shape control lie in the kernel of the nonholonomic constraint map. Consequently, the total nonholonomic momentum can be decomposed into the coupled nonholonomic momentum, $\genmom$,\footnote{For simplicity, we refer to it as nonholonomic momentum in the rest of this paper.} and the shape momentum, $\basemom$,
\begin{eqalign}
    \bar{\nonhmap} &= \begin{bmatrix}
    \multirow{2}{*}{$\nonhmap$} & 0 \\
    & \matrixid
    \end{bmatrix} \\
    \bar{\genmom} &= \begin{bmatrix}
        \genmom & \basemom
    \end{bmatrix} \\
    \genmom &= \left\langle\begin{bmatrix}
        \bodymom & \basemom
    \end{bmatrix}; \nonhmap\right\rangle.
\end{eqalign}
By substituting the decomposition, the kinematics is rearranged into the reconstruction equation, which expresses the system's body velocity as a function of the current nonholonomic momentum and the shape velocity control:
\begin{eqalign}
    \fibercirc &= \inv{\begin{bmatrix}
        \transpose{\nonhmap}\begin{bmatrix}
            \Inertiametric_{\fiber\fiber} \\ \Inertiametric_{\base\fiber}
        \end{bmatrix} \\
        \nonholonomic_{\fiber}
    \end{bmatrix}}\begin{bmatrix}
        \transpose{\genmom} - \transpose{\nonhmap}\begin{bmatrix}
            \Inertiametric_{\fiber\base} \\ \Inertiametric_{\base\base}
        \end{bmatrix}\basedot \\
        -\nonholonomic_{\base}\basedot
    \end{bmatrix} \\
    &= \mixedconn(\base)\basedot + \genmomconn(\base)\transpose{\genmom}, \label{eq:reconstruction}
\end{eqalign}
where $\mixedconn$ is commonly referred to as the system's local connection, representing kinematic effects, while $\genmomconn$ denotes the connection associated with nonholonomic momentum, incorporating dynamic effects. 

Note that systems with sufficiently many nonholonomic constraints to fully determine the body velocity from the controlled shape will have no nonholonomic momentum, i.e.,  
\begin{subequations} \label{eq:primarily_kinematic}
\begin{align}
    \fibercirc &= \mixedconn(\base)\basedot \\
    \genmom &\text{ is not defined}.
\end{align}
\end{subequations}
Additionally, systems whose dynamics are fully governed by momentum conservation, starting from rest, will take the form:  
\begin{subequations} \label{eq:purely_mechanical}
\begin{align}
    \fibercirc &= \mixedconn(\base)\basedot \\
    \genmom &= \dot{\genmom} = 0.
\end{align}
\end{subequations}
In these two cases, the system reduces to a principally kinematic system and a purely mechanical system, respectively \cite{shammas2007geometric}. Extensions to purely mechanical systems starting from nonzero momentum are discussed in \cite{yang2023geometric}.

\subsection{Momentum Evolution}

The momentum evolution can be derived from constrained Lagrangian mechanics:  
\beq
    \begin{bmatrix}
        \bodymomdot & \basemomdot
    \end{bmatrix} = \lagrangemult\nonholonomic + \begin{bmatrix}
        -\coadji_{\fibercirc}\bodymom + \coforce \\[0.75em]
        \parderiv{\lagrangian}{\base} + \lagrangeforce
    \end{bmatrix}, \label{eq:lagrangemechanics}
\eeq
where $\lagrangemult$ represents the constraint force due to the nonholonomic constraints, $\coadji_{\fibercirc}\bodymom$ denotes the rate of change of a constant momentum expressed in a moving body frame relative to a fixed observer in space, and $\coforce$ is the body force acting on the system's body frame. This body force includes viscous friction on the wheels of the roller racer and snakeboard, as well as nonlinear fluid drag forces on the swimmer at intermediate Reynolds numbers, as considered in this paper.\footnote{For force-controlled systems---such as those using jet thrusters attached to the body---this model can be generalized to include the controlled force in~$\coforce$.} The viscous friction is modeled as being linear in the wheel velocity, whereas the fluid drag force increases quadratically with the flow velocity:  
\beq
    \coforce =
    \begin{cases}
        -\sum_{i}\coAdj_{\inv{\fiber}_{i}}\transpose{\fibercirc}_{i}\dragmetric_{\text{friction}, i} & \text{for viscous friction} \\
        -\sum_{i}\coAdj_{\inv{\fiber}_{i}}\transpose{(\lvert\fibercirc_{i}\rvert\odot\fibercirc_{i})}\dragmetric_{\text{drag}, i}  & \text{for fluid drag},
    \end{cases}
\eeq
where $\odot$ denotes the element-wise product, $\lvert\bullet\rvert$ represents the element-wise absolute value, $\fiber_{i}$ is the position of the $i$-th component relative to the system's body frame, $\dragmetric_{\text{friction}, i}$ and $\dragmetric_{\text{drag}, i}$ are the viscous friction and fluid drag coefficients of the $i$-th component, expressed in its body frame, and $\Adj^{*}_{\bullet}$ denotes the dual adjoint matrix corresponding to the transformation.\footnote{In colloquial terms, $\Adj_{\bullet}$ and $\Adj^{*}_{\bullet}$ refer to the adjoint and dual adjoint operators that combine the cross-product operation---which converts linear and angular velocities or forces/momenta between different frames---with the rotation operation that expresses velocities or forces/momenta in body-aligned coordinates.} 

Similarly, $\lagrangeforce$ represents the torque acting on the joints, which includes contributions from viscous friction or quadratic fluid drag forces, as well as actuator-applied torques for shape control: 

\beq
    \lagrangeforce_{j} =
    \begin{cases}
        \bar{\lagrangeforce}_{j} - \sum_{i}\Adj^{*\theta}_{\fiber_{ij}}\transpose{\fibercirc}_{i}\dragmetric_{\text{friction}, i} & \text{for viscous friction} \\
        \bar{\lagrangeforce}_{j} - \sum_{i}\Adj^{*\theta}_{\fiber_{ij}}\transpose{(\lvert\fibercirc_{i}\rvert\odot\fibercirc_{i})}\dragmetric_{\text{drag}, i}  & \text{for fluid drag},
    \end{cases}
\eeq
where $\fiber_{ij}$ is the transformation matrix from the $i$-th link to the $j$-th joint, $\Adj^{*\theta}_{\bullet}$ denotes the row corresponding to rotation in the dual adjoint matrix, and $\bar{\lagrangeforce}_{j}$ represents the actuator torque.

The time derivative of the nonholonomic momentum can be computed by differentiating its definition:
\beq
    \genmomdot &= \left\langle\begin{bmatrix}
        \bodymomdot & \basemomdot
    \end{bmatrix}; \nonhmap\right\rangle + \left\langle\begin{bmatrix}
        \bodymom & \basemom
    \end{bmatrix}; \dot{\nonhmap}\right\rangle. \label{eq:momevo}
\eeq
Because $\nonhmap$ lies in the kernel of the nonholonomic constraint map, the inner product eliminates the constraint forces in~\eqref{eq:lagrangemechanics}. 

Finally, substituting the time derivative of the nonholonomic momentum along with the controlled position, velocity, and acceleration of the shape into~\eqref{eq:lagrangemechanics} allows for the calculation of the joint actuator torque $\lagrangeforce$.

\subsection{Examples}

The key features of the three examples of kinodynamic systems considered in this paper are summarized in Fig.~\ref{fig:systems}. Both the roller racer and the snakeboard are kinodynamic systems with one degree of freedom (DOF) of nonholonomic momentum, resulting from nonholonomic constraints and dynamic effects caused by internal relative motion between multiple bodies. For these systems, wheel friction is modeled as linear viscous friction \cite{ostrowski1998reduced, krishnaprasad2001oscillations}. 

The primary difference between the roller racer and the snakeboard lies in their actuation. The roller racer has a single actuated DOF---the steering angle---where the inertia rearrangement and constraint directions are coupled to the steering control. In contrast, the snakeboard features two actuated DOFs: one for the rotor and the other for the trucks’ steering (with the trucks assumed to steer together in opposite directions). This configuration decouples the inertia rearrangement from the constraint directions.

Unlike the roller racer and the snakeboard, the swimmer at intermediate Reynolds numbers does not involve direct nonholonomic constraints. Instead, its dynamics are modeled by Morison's equation, incorporating anisotropic fluid-added mass and hydrodynamic drag forces acting on each body \cite{brennen2006internet, boyer2010poincare, porez2014improved}. Consequently, the swimmer possesses three DOFs of nonholonomic momentum, corresponding to the position dimensions. In this case, the kinematic and dynamic effects are strongly coupled, with the fluid-added mass and drag forces being shape-dependent and directly acting on the joints.

\section{Variational Gait Optimization Based on Lie Group Integrator}

In this section, we introduce the variational gait optimization method for kinodynamic systems. The Lie group integrator provides an efficient and accurate method to compute the gradients and Jacobians of the gait-induced displacement with respect to the gait parameters, enabling effective variational gait optimization. We begin by defining the displacement, nonholonomic momentum, and power consumption corresponding to a gait cycle. These terms are fundamental concepts in the gait optimization problem, as most objective and constraint functions rely on them. Once the gradients of these terms are computed, the chain rule can be applied to derive the gradients of the optimization objective and constraint functions, thereby achieving variational gait optimization.

Without loss of generality, consider a gait $\gait$ in the shape space, parameterized by the gait parameters $\varparam$. The displacement induced by this gait is given by  
\beq
    \gaitdisp = \Prodi_{0}^{\period}\left(\matrixid+\fibercirc(\vartime) d\vartime\right) \label{eq:fwdeuler},
\eeq
where $\period$ represents the gait period, and $\prodi$ is the product integral. This formulation essentially is a Lie group integrator.  

Additionally, as shown in~\eqref{eq:momevo}, the nonholonomic momentum over the gait period is given by the solution to the ordinary differential equation for momentum evolution:  
\beq
    \genmom(\vartime) = \genmom(0) + \int_{0}^{\vartime}\genmomdot\bigl((\base(\altvartime), \basedot(\altvartime), \genmom(\altvartime)\bigr)d\altvartime, \label{eq:momevoint}
\eeq 
where $\genmom(0)$ is the initial nonholonomic momentum at the start of the gait execution.  

Finally, the power consumption of executing the gait is defined as the integral of the squared actuator torque $\|\lagrangeforce(\vartime)\|^{2}_{2}$ over the gait period:  
\beq  
    \avgenergy = \int^{\period}_{0}\|\lagrangeforce(\vartime)\|^{2}_{2} d\vartime. \label{eq:energy}
\eeq   

\subsection{Gradient of Gait-Induced Displacement}

\begin{figure}[!t]
\centering
\includegraphics[width=0.9\linewidth]{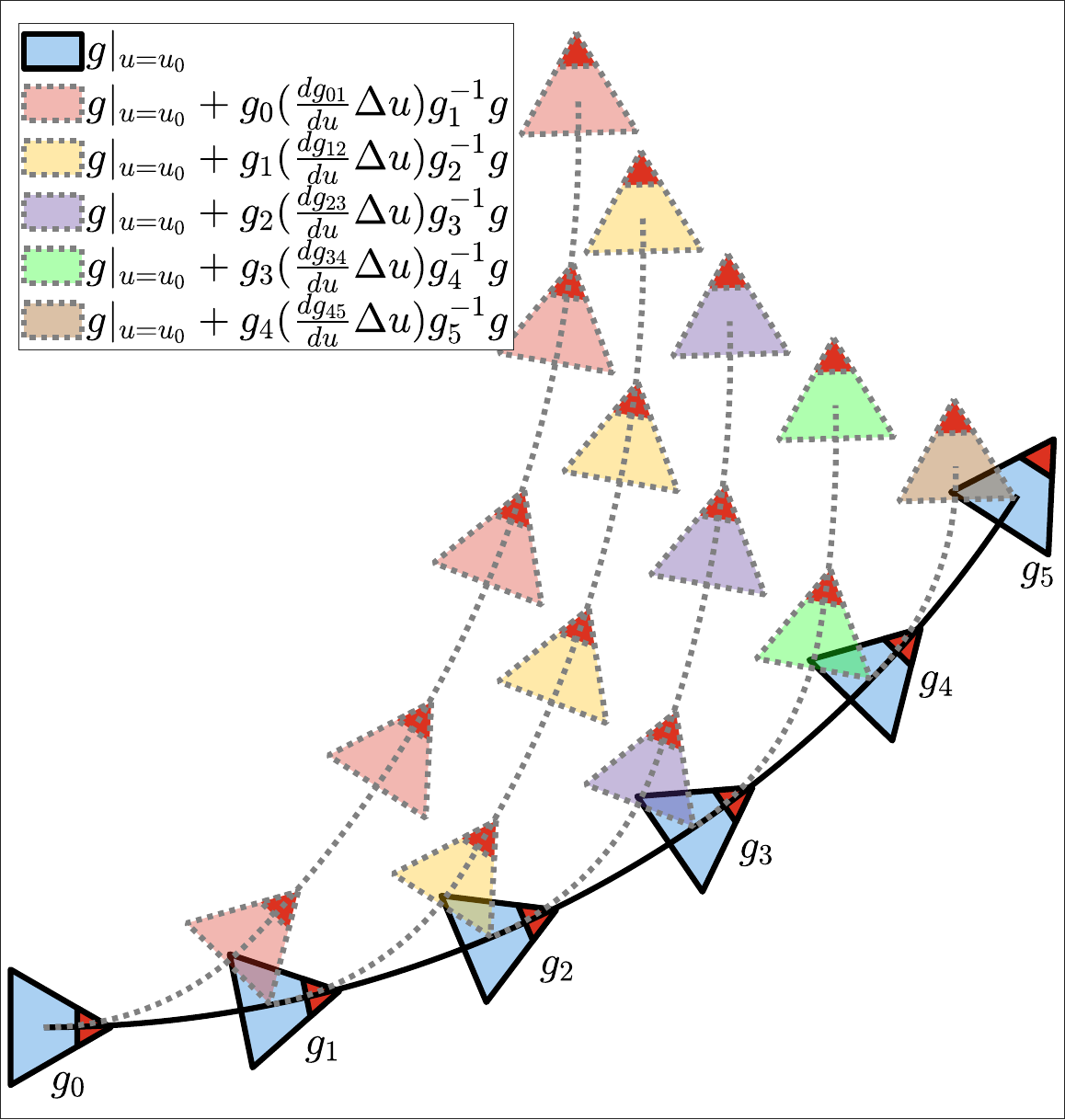}
\caption{Demonstration of gradient computation for variational gait optimization based on Lie group integrator. For a principally kinematic system, $\fibercirc = \begin{bmatrix} 1 & 0 & u \end{bmatrix}^{\intercal}$, the position trajectory $g\vert_{u=u_0}$ is represented by the solid line. The variation in the trajectory due to a control change $\Delta u$ is computed by summing the gradient contributions from the control change at each segment along the trajectory. The contribution of each segment is depicted as the dashed trajectory. Due to group symmetries, a control change at any segment does not affect the shape of the remaining trajectory, similar to rotating the joints of a robotic arm.}
\label{fig:gradient}
\end{figure}

By differentiating the Lie group integral with respect to the gait parameters $\varparam$, we obtain
\begin{eqalign}
    \parderiv{\gaitdisp}{\varparam} &= \int_{0}^{\period} \fiber(\vartime)\left(\parderiv{\fibercirc(\vartime)}{\varparam} + \frac{\fibercirc(\vartime)}{\period}\parderiv{\period}{\varparam}\right)\inv{\fiber}(\vartime)\gaitdisp d\vartime \\
    &= \int_{0}^{\period} \Adj_{\fiber(\vartime)}\left(\parderiv{\fibercirc(\vartime)}{\varparam} + \frac{\fibercirc(\vartime)}{\period}\parderiv{\period}{\varparam}\right)\gaitdisp d\vartime, \label{eq:disgrad}
\end{eqalign}
The second term in the sum inside the parentheses arises because the period $\period$ is treated as a function of the gait parameters $\varparam$, requiring consideration of the integration bounds during differentiation.

Using the reconstruction equation in~\eqref{eq:reconstruction}, the derivative of the system's body velocity can be expressed as
\beq
    \parderiv{\fibercirc(\vartime)}{\varparam} = \parderiv{\fibercirc_{\vartime}}{\base_{\vartime}}\parderiv{\base_{\vartime}}{\varparam} + \parderiv{\fibercirc_{\vartime}}{\basedot_{\vartime}}\parderiv{\basedot_{\vartime}}{\varparam} + \parderiv{\fibercirc_{\vartime}}{\genmom_{\vartime}}\parderiv{\genmom_{\vartime}}{\varparam},
\eeq
where time-varying variables $[\bullet](\vartime)$ are denoted as $[\bullet]_{\vartime}$ for clarity in the equations.
Because the gait in the shape space is parameterized by $\varparam$, the derivatives of the shape position and velocity can be directly obtained from the parameterization function. In contrast, the derivative of the momentum is more complex, because it involves the momentum evolution, which will be addressed in the next section.

\subsection{Gradient of Nonholonomic Momentum}

Based on the nonholonomic momentum solution in~\eqref{eq:momevoint}, the chain rule can be applied to compute the derivative of the nonholonomic momentum:  
\beq \label{eq:linear_system}
    \parderiv{\genmom(\vartime)}{\varparam} &= \int^{t}_{0}\left(\parderiv{\genmomdot_{\altvartime}}{\base_{\altvartime}}\parderiv{\base_{\altvartime}}{\varparam} + \parderiv{\genmomdot_{\altvartime}}{\basedot_{\altvartime}}\parderiv{\basedot_{\altvartime}}{\varparam} + \parderiv{\genmomdot_{\altvartime}}{\genmom_{\altvartime}}\parderiv{\genmom_{\altvartime}}{\varparam}\right)d\altvartime \\
    &\quad+\frac{\genmom(\vartime) - \genmom(0)}{\period}\parderiv{\period}{\varparam},
\eeq 
where, as before, time-varying variables $[\bullet](\altvartime)$ are denoted as $[\bullet]_{\altvartime}$ for clarity in the equations. The last term arises because the period $\period$ depends on the gait parameters $\varparam$, requiring consideration of the integration bounds during differentiation.

Similarly, the gradient of the energy consumption in~\eqref{eq:energy} for executing the gait can be derived in the same manner, as it follows a similar sum-integral structure.

A key difference between the gradient of momentum and that of displacement or energy consumption is the temporal autocorrelation of momentum. Specifically, the momentum at time $\vartime$ depends on the time series of momentum preceding $\vartime$, because $\genmomdot$ is a function of $\genmom$ itself. This temporal dependence transforms the gradient computation into solving a linear system after discretization. Consequently, this distinction affects the computational accuracy of the gradients post-discretization, which we will discuss in the next section.

\subsection{Invariance and Gradient Error Analysis}

For principally kinematic and purely mechanical systems described by~\eqref{eq:primarily_kinematic} and~\eqref{eq:purely_mechanical}, which do not involve nonholonomic momentum, the gradients of gait-induced displacement are invariant. Specifically, as illustrated in Fig.~\ref{fig:gradient}, the body velocities are trajectory-independent, meaning any variation in the body velocity at a point on the trajectory applies only a left group action to the subsequent trajectory. Integrating this variation along the trajectory, as in~\eqref{eq:disgrad}, yields the corresponding variation in the final displacement. Consequently, once the trajectory is simulated, the displacement gradient along it can be directly computed, with accuracy linearly proportional to the discretization density. These characteristics allow for an efficient and accurate gradient-based variational gait optimization across a wide range of principally kinematic and purely mechanical systems.

In contrast, for the kinodynamic systems considered in this paper, the presence of nonholonomic momentum breaks this invariance. The momentum evolution is a self-dependent ordinary differential equation, causing momentum gradient errors to scale quadratically with discretization density. Because the displacement gradient depends on the momentum gradient, these errors propagate into the displacement gradient as well. In this case, optimizing gaits or trajectories over long horizons becomes impractical, because discretization errors or model mismatches accumulate. Instead, we focus on optimizing periodic gaits and short-term transitions between them. These motion primitives are subsequently combined to construct complete motion plans, as detailed in the motion planning section below.

\section{Geometric Motion Planning}

Motion planning for kinodynamic systems is inherently more complex than for principally kinematic or purely mechanical systems due to their time-dependent dynamics. In kinematic systems, where dynamics are quasi-static, motion planning can often be achieved using simple gaits to locomote directly in the desired direction. However, in kinodynamic systems, the presence of nonholonomic momentum requires accounting for the evolution of momentum, adding additional complexity to the planning process. Efficient locomotion involves first accumulating momentum and then directing it to drive towards the desired direction. This principle appears in many real-world examples, such as roller racer or snakeboard riders swinging to accelerate before steering towards their target, or fish stroking to generate thrust and then gliding towards the desired direction using their streamlined bodies. In this paper, we leverage the gradient computations based on the Lie group integrator discussed earlier to identify gaits and transitions, enabling the construction of comprehensive motion plans for kinodynamic systems.  

We begin by considering the following optimization problem, which maximizes the linear displacement over a time horizon $\period_{\infty}$, significantly longer than a single gait period $\period$ (i.e., $\period_{\infty} \gg \period$), subject to joint position limits and energy consumption constraints:  
\begin{eqalign}
    \max &\quad \left\|\fiber^{[x, y]}(\period_{\infty})\right\|_{2} \\
    s.t. &\quad \base(\vartime) \in \mathcal{\basespace}, \quad \forall t \in [0, \period_{\infty}] \\
    &\quad \|\lagrangeforce(\vartime)\|^{2}_{2} < \constraint, \quad \forall \vartime \in [0, \period_{\infty}],
\end{eqalign} 
where $\fiber^{[x, y]}(\period_{\infty})$ denotes the linear displacement over the horizon $\period_{\infty}$, $\mathcal{\basespace}$ represents the range of motion of the robot joints, and $\constraint$ defines the power consumption limit. Due to energy dissipation from velocity-dependent friction or fluid drag, as well as constraints on energy consumption and the range of motion of the joints, the robot will eventually converge to a steady state with periodic motion and state. This allows us to decompose the original problem into four sub-problems, solving for multiple gaits and their transitions to construct a comprehensive motion plan, as illustrated in Fig.~\ref{fig:roller_racer_gait}:
\begin{enumerate}
    \item Steady-state gait optimization: Finding the optimal gait $\gait_{\steadystate}$ and periodic nonholonomic momentum state $\genmom_{\steadystate}$ that maximize the steady-state speed.  
    \item Acceleration gait optimization: Identifying the acceleration gait $\gait_{\acceleration}$ that optimally accelerates the robot to the steady state.  
    \item Transition gait optimization: Determining the transition gaits $\gait_{\transition}$ for:
    \begin{enumerate}
        \item Starting from rest at the nominal shape and transitioning to the acceleration gait,
        \item Transitioning from the acceleration gait to the steady-state gait,
        \item Returning from the steady-state gait to the nominal shape.  
    \end{enumerate}
    \item Turning gait optimization: Finding the turning gait $\gait_{\turning}$ to redirect the steady-state robot to a new direction.  
\end{enumerate}

\subsection{Steady-State Gait Optimization}

We first address steady-state gait optimization, which involves maintaining periodic motion and state while maximizing velocity and satisfying specific constraints. This gait optimization can be formulated as:
\begin{eqalign}
    \max_{\gait_{\steadystate}, \genmom_{\steadystate}} &\quad \frac{1}{\period_{\steadystate}}\left\|\fiber_{\gait_{\steadystate}}^{[x, y]}\right\|_{2} \\
    s.t. &\quad \genmom_{\steadystate}(0) = \genmom_{\steadystate}(\period_{\steadystate}) \label{eq:momconstr} \\
    &\quad \fiber_{\gait_{\steadystate}}^{\theta} = 0 \label{eq:orientconstr} \\ 
    &\quad \base_{\steadystate}(t) \in \mathcal{\basespace}, \quad \forall \vartime \in [0, \period_{\steadystate}] \label{eq:jointconstr} \\ 
    &\quad \frac{\avgenergy_{\steadystate}}{\period_{\steadystate}} < \constraint \label{eq:energyconstr} \\ 
    &\quad \gait_{\steadystate}, \genmom_{\steadystate} \text{ are feasible}. \label{eq:feasibility}
\end{eqalign}
In this formulation, the objective function seeks to maximize the average translational velocity of the gait. Constraint \eqref{eq:momconstr} ensures that the nonholonomic momentum $\genmom_{\steadystate}$ is periodic throughout the gait cycle. Constraint \eqref{eq:orientconstr} requires the robot to maintain its orientation before and after the gait cycle. Constraint \eqref{eq:jointconstr} enforces that joint positions remain within their prescribed limits. Constraint \eqref{eq:energyconstr} requires that the average energy consumption stays within the allowable range. Finally, \eqref{eq:feasibility} serves as an auxiliary feasibility condition, ensuring that the steady state can be reached from rest via an acceleration gait while adhering to the power consumption constraint. This avoids potential energy barriers that could render the steady state inaccessible. Details on this auxiliary feasibility problem, as well as the design of the acceleration gait, will be discussed in the next section.

\subsection{Acceleration Gait Optimization}

\begin{figure}[!t]
\centering
\includegraphics[width=\linewidth]{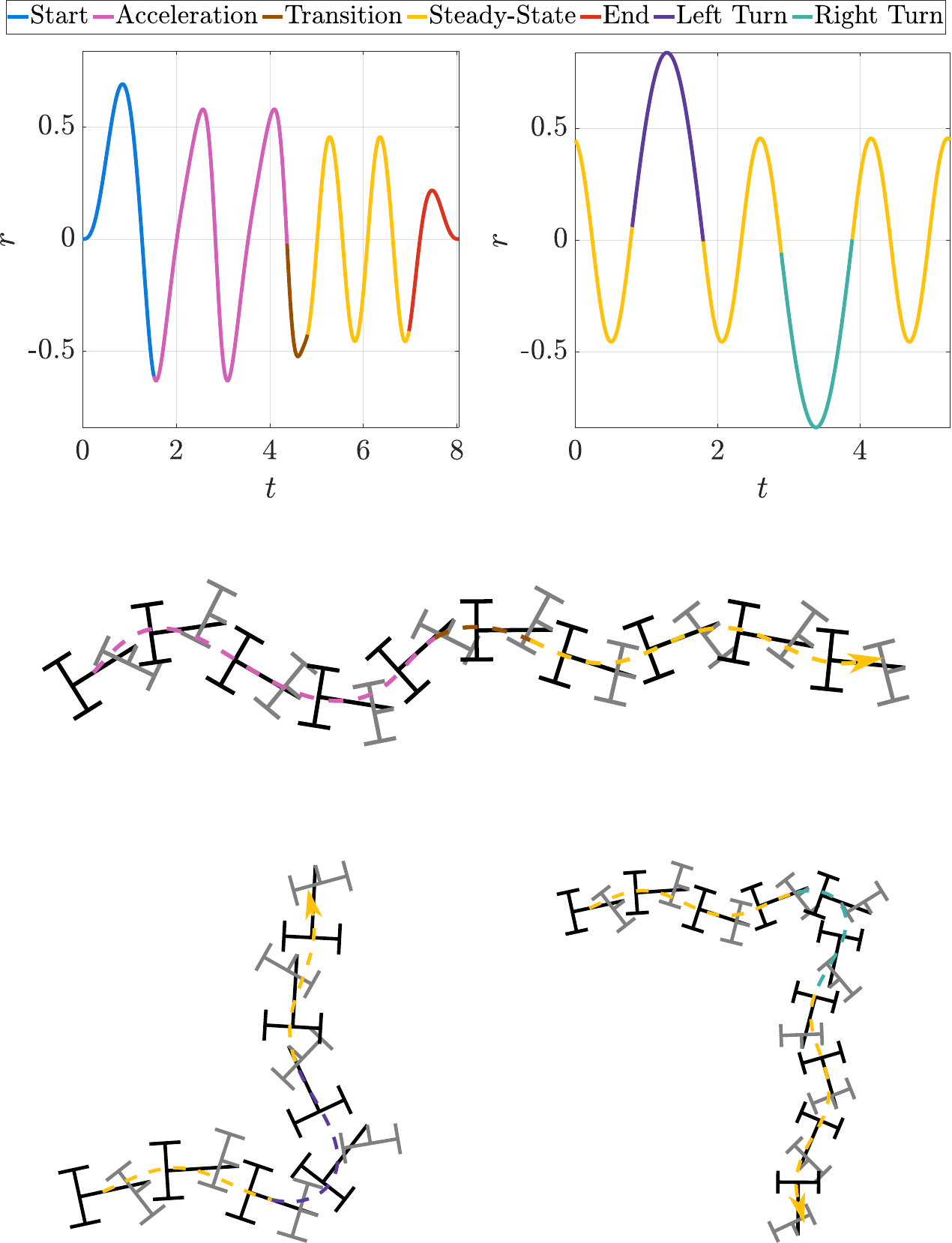}
\caption{Illustration of the optimal gaits and transitions for the roller racer (top) and their demonstration on the robot (middle: acceleration-transition-steady-state gait; bottom: steady-state-left-turn-steady-state and steady-state-right-turn-steady-state gaits). In the demonstration, the dashed line represents the robot’s position trajectory, with its color indicating the corresponding gait being executed at each point.}
\label{fig:roller_racer_gait}
\end{figure}

Due to the time-varying dynamics of kinodynamic systems, naively executing the steady-state gait for acceleration is impractical, because it may violate power constraints, drift in direction, or require excessive convergence time. To address this, we explore effective acceleration gaits to reach the steady state in this section.

We begin by introducing the feasibility problem formulated in~\eqref{eq:feasibility}. The objective is to find an acceleration gait that accelerates the system from rest to a steady state, achieving the average kinetic energy level similar to that of the optimal steady-state gait without violating joint or power constraints:
\begin{eqalign}
    \underset{\gait_{\feasibility}, \genmom_{\feasibility}}{\mathrm{find}} & \\
    s.t. &\quad \genmom_{\feasibility}(0) = \genmom_{\feasibility}(\period_{\feasibility}) \\
    &\quad \fiber_{\gait_{\feasibility}}^{\theta} = 0 \\
    &\quad \base_{\feasibility}(t) \in \mathcal{\basespace}, \quad \forall \vartime \in [0, \period_{\feasibility}] \\
    &\quad \frac{\avgenergy_{\feasibility, 0}}{\period_{\feasibility}} < \constraint ,\quad \frac{\avgenergy_{\feasibility}}{\period_{\feasibility}} < \constraint \\
    &\quad \overline{\kineticenergy}_{\feasibility} > \overline{\kineticenergy}_{\steadystate},
\end{eqalign}
where $\avgenergy_{\feasibility, 0}$ represents the power consumption during the first gait cycle starting from rest (i.e., zero nonholonomic momentum),\footnote{Technically, the average power consumption is not a convex function of the nonholonomic momentum, which implies that enforcing constraints only at zero momentum and steady-state momentum does not guarantee compliance with the power limit at intermediate points. However, for the kinodynamic systems studied in this paper, the power consumption peak typically occurs either at the start or in the steady state. In practice, one can validate the power consumption limits after optimizing the gait. Addressing the power consumption constraints more rigorously during acceleration remains an open challenge, which we leave for future work.} and $\overline{\kineticenergy}_{\steadystate}$ and $\overline{\kineticenergy}_{\feasibility}$ denote the average kinetic energy of the steady-state gait and that of the feasible acceleration gait in its steady state, respectively. This feasibility problem ensures that an acceleration gait exists to bring the system from rest to the kinetic energy level of the steady-state gait while respecting all constraints, thereby enabling a further transition to the steady-state gait. In the steady-state gait optimization, this feasibility problem is solved as an auxiliary problem alongside to ensure the feasibility of the steady-state gait.

For the actual acceleration gait optimization, simply ensuring the existence of feasible gaits is insufficient, because we also want to maximize efficiency and maintain the heading angle. To this end, we propose an approach for optimizing acceleration gaits. A key characteristic of the acceleration behavior in the kinodynamic systems studied in this paper is that, during the transition from a stationary state to a steady state, the average translational velocity of each gait cycle, $\bar{\fiberdot}^{[x, y]}$, converges to the steady-state velocity, $\bar{\fiberdot}^{[x, y]}_{\steadystate}$, in a manner analogous to a linear mass-damper system:  
\beq  
    \bar{\fiberdot}^{[x, y]}(t) \approx (1 - e^{-kt})\bar{\fiberdot}^{[x, y]}_{\steadystate},  
\eeq  
where $k$ represents the convergence rate or equivalent damping coefficient. This behavior emerges naturally from the dynamic model: executing repeated acceleration gaits continuously adds energy to the system, while energy dissipation due to viscous friction is proportional to velocity.  

For systems like the roller racer and snakeboard, which assume linear viscous friction, this analogy is particularly appropriate, as their wheels consistently generate friction through continuous rolling. For intermediate Reynolds number swimmers, despite the fluid-added mass being acceleration-dependent and the fluid drag quadratically velocity-dependent, they are anisotropic and shape-dependent. Consequently, an efficient acceleration gait minimizes the equivalent total fluid drag during acceleration. Thus, the linear mass-damper model remains a valid approximation for these systems, as illustrated in the following figure.

Using the linear mass-damper approximation model for the acceleration period, we can reformulate the feasibility problem as an acceleration gait optimization problem:
\begin{eqalign}
    \min_{\gait_{\acceleration}, \genmom_{\acceleration}} &\quad \left\|\bar{\fiberdot}^{[x, y]}_{\steadystate}\right\|\period_{\transient} - \fiber_{\transient} \\
    s.t. &\quad \genmom_{\acceleration}(0) = \genmom_{\acceleration}(\period_{\acceleration}) \\
    &\quad \fiber_{\gait_{\acceleration}, 0}^{\theta} = 0 ,\quad \fiber_{\gait_{\acceleration}}^{\theta} = 0 \\
    &\quad \frac{\avgenergy_{\acceleration, 0}}{\period_{\acceleration}} < \constraint ,\quad \frac{\avgenergy_{\acceleration}}{\period_{\acceleration}} < \constraint \\
    &\quad \base_{\acceleration}(t) \in \mathcal{\basespace}, \quad \forall \vartime \in [0, \period_{\acceleration}] \\
    &\quad \overline{\kineticenergy}_{\acceleration} > \overline{\kineticenergy}_{\steadystate},
\end{eqalign}
where the subscript $[\bullet]_{\transient}$ denotes the total transient period, which we define as the rise time required to reach 90\% of the steady-state velocity, calculated based on the linear mass-damper model. The objective function seeks to determine the optimal acceleration gait that minimizes the deviation from the steady-state velocity during the transition. The term $\fiber_{\gait_{\acceleration}, 0}^{\theta}$ represents the orientation difference before and after the first gait cycle starting from rest. Constraints on this term, combined with constraints on the orientation difference at the steady state, ensure a consistent heading angle throughout the acceleration period.\footnote{Unlike the non-convex power consumption discussed earlier, the orientation displacement before and after each gait cycle is a linear function of the momentum change. This relationship arises because the body velocity is computed by applying the linear mapping $\genmomconn$ to the momentum, as shown in~\eqref{eq:reconstruction}. Because $\genmomconn$ is shape-dependent and the shape remains consistent across each cycle for shape-controlled gaits, the orientation change produced during each gait cycle is linearly related to the nonholonomic momentum. Consequently, by enforcing zero orientation displacement at the beginning and at the steady state, the orientation remains unchanged throughout the acceleration period.}

\subsection{Transition Gait Optimization}

With the optimal gaits derived in the previous section, a complete motion plan can be constructed by combining these gaits. The remaining challenge is to enable smooth transitions between these gaits without loss of generality. In this section, we propose a method for gait transitions using polynomials.  

To ensure a smooth transition while respecting power consumption constraints, we employ a fifth-order polynomial to connect between the two gaits. This polynomial guarantees continuity up to the shape acceleration at both ends. Because the initial and final shape positions, velocities, and accelerations are fully determined by the gaits at either end of the transition, all polynomial coefficients can be computed directly. The transition gait optimization problem for determining the transition gait $\gait_{\transition}(\vartime_{0}, \vartime_{1}, \period_{\transition})$ for transitioning from gait $(\gait_{0}, \genmom_{0})$ to gait $(\gait_{1}, \genmom_{1})$ is formulated as:
\begin{eqalign}
    \min_{\vartime_{0}, \vartime_{1}, \period_{\transition}} &\quad \left\|\left(\fiber_{\gait_{\transition}}^{\theta} + \fiber_{\gait_{0}}^{\theta}(t_{0}) - \operatorname{atan2}(\fiber_{\gait_{0}}^{y}, \fiber_{\gait_{0}}^{x})\right)\right. \nonumber \\
    &\quad\quad- \left.\left(\fiber_{\gait_{1}}^{\theta}(t_{1}) - \operatorname{atan2}(\fiber_{\gait_{1}}^{y}, \fiber_{\gait_{1}}^{x})\right)\right\|_{2} \\
    s.t. &\quad \gait_{\transition}^{(n)}(0) = \gait_{0}^{(n)}(\vartime_0), \quad n = 0, 1, 2 \label{eq:polyconstr1} \\
    &\quad \gait_{\transition}^{(n)}(\period_{\transition}) = \gait_{1}^{(n)}(\vartime_1), \quad n = 0, 1, 2 \label{eq:polyconstr2} \\
    &\quad \genmom_{0}(t_0) + \genmom_{\transition}(T_{\transition}) = \genmom_{1}(t_1) \label{eq:polymomconstr} \\
    &\quad \base_{\transition}(\vartime) \in \mathcal{\basespace}, \quad \forall \vartime \in [0, \period_{\transition}] \label{eq:polyjointconstr} \\
    &\quad \frac{\avgenergy_{\transition}}{\period_{\transition}} < \constraint, \label{eq:polyenergyconstr}
\end{eqalign}
where $t_{0}$ and $t_{1}$ are the phases of the two gaits connected by the polynomial, $\period_{\transition}$ represents the transition duration, and $\genmom_{\transition}$ is the change in momentum relative to the initial momentum at the beginning of the transition. The objective function minimizes orientation drift during the transition, ensuring consistency in the translational motion direction between the two gaits while accounting for orientation oscillations within each gait. Constraints \eqref{eq:polyconstr1} and \eqref{eq:polyconstr2} define the 5th-order polynomial coefficients based on the shape position, velocity, and acceleration of the gaits before and after the transition. Constraint \eqref{eq:polymomconstr} enforces momentum continuity between the two gaits. Constraints \eqref{eq:polyjointconstr} and \eqref{eq:polyenergyconstr} address joint limits and power consumption. During execution, the transition is triggered once the acceleration gait reaches the energy level of the steady-state gait.  

The same approach can be adapted for transitions at the start or end of motion. To initiate acceleration from rest at the nominal shape, we fix $\gait_{0}(t_{0})$ and $\genmom_{0}$ to the initial state. Conversely, to transition from a steady-state gait to gliding, $\gait_{1}(t_{1})$ is fixed to the gliding shape, and the momentum constraint is relaxed.

\begin{figure}[!t]
\centering
\includegraphics[width=\linewidth]{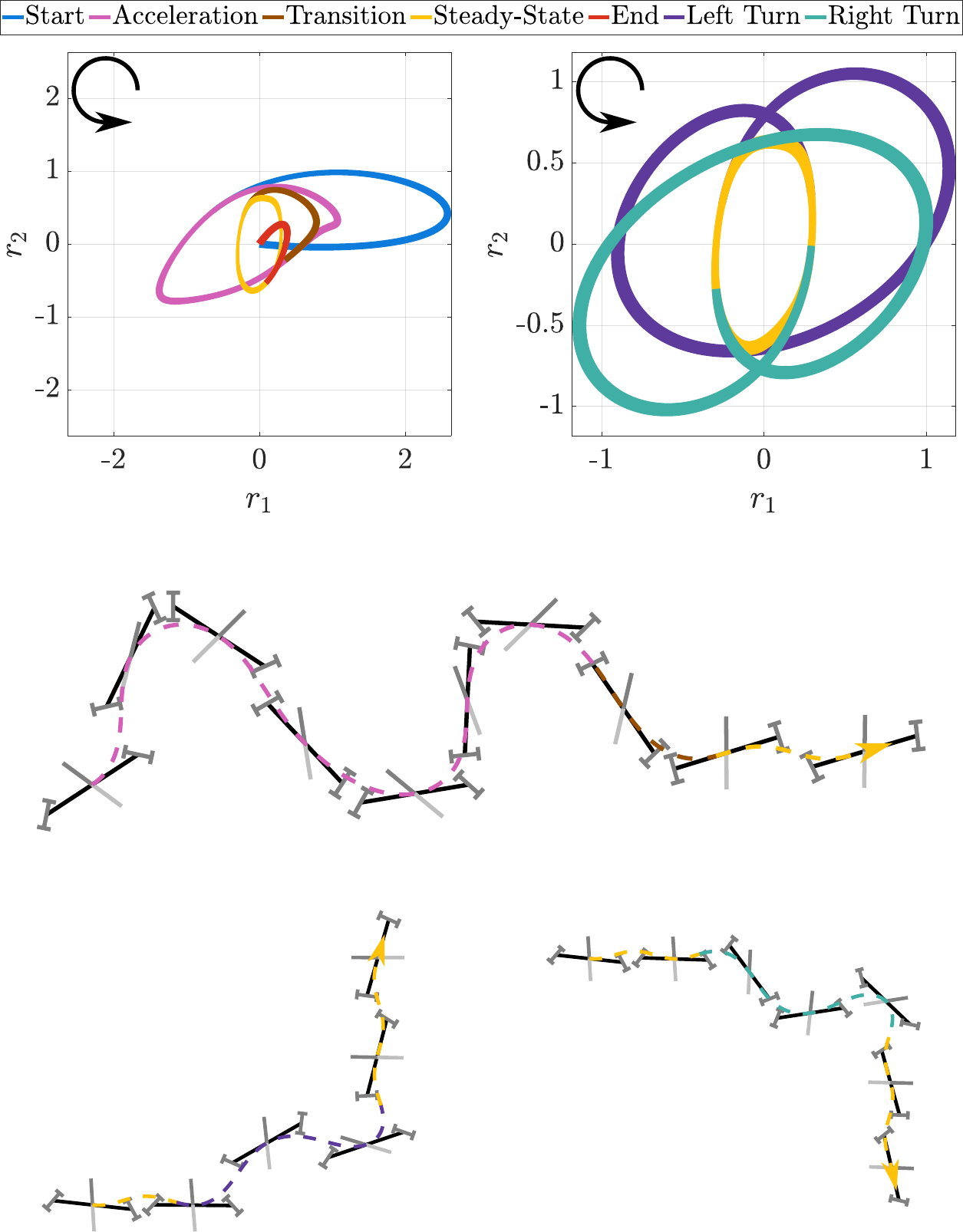}
\caption{Illustration of the optimal gaits and transitions for the snakeboard (top) and their demonstration on the robot (middle: acceleration-transition-steady-state gait; bottom: steady-state-left-turn-steady-state and steady-state-right-turn-steady-state gaits). The thickness of the gait cycles indicates the pacing of the gait, with thick lines representing slower shape changes and thin lines representing faster changes. Black arcs with arrows denote the direction of the gait. In the demonstration, the dashed line represents the robot’s position trajectory, with its color indicating the corresponding gait being executed at each point.}
\label{fig:snakeboard_gait}
\end{figure}

\subsection{Turning Gait Optimization}

In the previous section, we discussed gait optimization and motion planning for maximizing translational velocity in a single direction, but did not address motion in other directions. As demonstrated in \cite{ostrowski1995mechanics, krishnaprasad2001oscillations, morgansen2007geometric}, these kinodynamic systems exhibit small-time local controllability in all three directions, enabling the design of gaits and motion plans for lateral motion or rotation using the same approach. 

However, in practice, roller racer and snakeboard riders rarely move sideways or rotate on the spot, and similarly, fish seldom swim sideways. Instead, they typically accelerate to accumulate momentum and then use this momentum to turn toward the desired direction. Therefore, in this section, we extend the transition gait optimization to turning motions, thereby extending the motion planning capabilities.

Turning gait optimization is introduced as an extension of transition gait optimization, where the turning gait, $\gait_{\turning}(\vartime_{0}, \vartime_{1}, \period_{\turning})$, is preceded and followed by the same steady-state gait and satisfies the requirement that the orientation change equals the desired turning angle:  
\begin{eqalign}
    \min_{\vartime_{0}, \vartime_{1}, \period_{\turning}} &\quad \left\|\fibercirc_{\gait_{\steadystate}}(\vartime_1)-\fibercirc_{\gait_{\turning}}(\period_{\turning})\right\|_{\Inertiametric_{\fiber\fiber}(\vartime_{1})} \\
    s.t. &\quad \gait_{\turning}^{(n)}(0) = \gait_{\steadystate}^{(n)}(\vartime_0), \quad n = 0, 1, 2 \\
    &\quad \gait_{\turning}^{(n)}(\period_{\turning}) = \gait_{\steadystate}^{(n)}(\vartime_1), \quad n = 0, 1, 2 \\
    &\quad \fiber_{\gait_{\turning}}^{\theta} + \fiber_{\gait_{\steadystate}}^{\theta}(t_{0}) - \fiber_{\gait_{\steadystate}}^{\theta}(t_{1}) = \Delta\theta \\
    &\quad \base_{\turning}(\vartime) \in \mathcal{\basespace}, \quad \forall \vartime \in [0, \period_{\turning}] \\
    &\quad \frac{\avgenergy_{\turning}}{\period_{\turning}} < \constraint,
\end{eqalign} 
where $\Delta\theta$ represents the desired turning angle, and the third constraint ensures that the directional change across the turning gait equals the turning angle. $\left\|\bullet\right\|_{\Inertiametric_{\fiber\fiber}(t_{1})}$ denotes the inner product weighted by the inertia matrix, evaluated at the steady-state gait shape at \( t_1 \). Because turning by gliding inevitably leads to energy loss, momentum continuity before and after the turn cannot be enforced. Instead, the objective function minimizes the difference in body velocity between the steady state and the post-turn state, weighted by the inertia matrix, effectively measuring the kinetic energy loss.

It is also possible to formulate a steering gait by enforcing zero energy loss and nonzero rotational displacement. A straightforward approach involves modifying the steady-state gait using the same transition formula---adding a segment to introduce rotational displacement without energy loss. However, this approach does not always yield feasible solutions, and the achievable steering angle is inherently limited.

\subsection{Implementation Details}

We solve the optimization using \texttt{fmincon} in MATLAB with the interior-point method, providing the cost and gradient from our method and approximating the Hessian via the Broyden-Fletcher-Goldfarb-Shanno (BFGS) algorithm. Our approach is also compatible with other nonlinear programming solvers.

As with any nonlinear optimization, the solution depends on the initial guess and may converge to a local minimum. For steady-state gait optimization, we choose a gait in which the shape variables oscillate sinusoidally out of phase and with an offset (tracing a circle in the shape space centered on the mean joint values) and its corresponding steady-state momentum as the initial guess to maintain the heading angle and respect joint limits. Then we seed the acceleration gait with the optimal steady-state gait. For transition, turning, start, and end gaits---parameterized by polynomials with fewer decision variables---we perform a grid search for initial guesses of the gait percentiles and initialize the period with the average of the adjacent gaits.

The computational complexity scales with the system dimension and discretization density. For high-dimensional systems, the main cost per iteration is solving the linear system in~\eqref{eq:linear_system}, which scales as $O(n^2)$. In our implementation, discretizing the gait cycle into 100 segments allows each gait's optimization to complete in less than one minute. Because gait optimization is performed offline and online execution only requires retrieving solutions from a pre-computed library, our method scales well to higher-dimensional systems.

\section{Simulation Results}

To evaluate the proposed approach, we apply it to all the three representative example systems to calculate a comprehensive motion plan involving start from the rest, accelerating to the steady-state energy level, transition between the acceleration and the steady-state gait, turning left and right, and smoothly end the run by resting at a nominal shape. For gait optimization, the steady-state and acceleration gaits are parameterized as fourth-order Fourier series.

\subsection{Roller Racer}

The roller racer is modeled based on the physical robot. The robot has a total weight of approximately 2.5 kg, with the rear cart weighing 1.7 kg and the front cart 0.8 kg. The distances from the joints to the center of mass are about 0.13 meters for the front cart and 0.05 meters for the rear cart. The wheels are spaced 0.07 meters apart laterally. The viscous friction coefficient was calculated based on the experimental data. The actuator force limit is set according to the hardware datasheet. The actuator range limits are set to avoid the center of mass extending beyond the supporting polygon.

Fig.~\ref{fig:roller_racer_gait} illustrates the optimal gaits and transitions for the roller racer and their demonstration. Because the system has only one actuated degree of freedom, most of the gaits resemble sinusoidal waves, except for the start and acceleration gaits, which feature different pacing on the rising and falling edges. These gaits are specifically designed to accelerate as quickly as possible while maintaining the heading angle under increasing momentum. The turning gait exhibits the largest wave magnitude to steer the front cart effectively, utilizing the accumulated momentum to achieve turns with minimal energy loss. Notably, the two turning gaits show symmetry in shape space. For all other gaits, the wave amplitude gradually decreases as momentum increases. This reflects a design principle where higher drive amplitudes are required for acceleration, while more compliant control is necessary to sustain high speeds in the steady state. The start, transition, and end gaits, ensure smooth transitions between the nominal zero shape and the cyclic gaits. These transitions are achieved using a half-period wave with a shape similar to the cyclic gaits being connected and an appropriate magnitude, facilitating heading angle maintenance and momentum matching.

The first row of Fig.~\ref{fig:simulation} illustrates the position trajectory and momentum history of the simulation results of a roller racer executing the comprehensive motion plan starting from rest. Because the motion plan was also designed for experimental implementation, the acceleration phase was reduced by one cycle from the duration required to reach 90\% of the steady-state velocity to accommodate environmental size constraints. Despite this adjustment, the resulting trajectory demonstrates promising performance, including starting the robot from the nominal zero shape, accelerating to the desired energy level, smoothly transitioning to the steady-state gait, performing left and right turns for a quarter corner, and finally gliding at the nominal zero shape. The momentum trajectory highlights key differences between the acceleration and steady-state cycles and illustrates the minimal energy loss during the turning phases.

\subsection{Snakeboard}

The snakeboard physics model is based on \cite{ostrowski1996mechanics}, with a total mass of 6 kg and inertias of 0.06, 0.167, and $\text{0.00167 kg}\cdot \text{m}^\text{2}$ for the body, rotor, and wheelsets, respectively. The board has a total length of 0.6 m, and the viscous friction coefficient is assumed to be 0.15. The range of rotor rotation is constrained to one cycle, and the range of wheel rotation mimics that of ankle joints.  

Fig.~\ref{fig:snakeboard_gait} illustrates the optimal gaits and transitions for the snakeboard in the shape space, along with their demonstration on the robot. Due to the limited range of both shapes, the gaits resemble circular cycles with varying radii and localized deformations. Unlike the roller racer, the acceleration and steady-state gaits for the snakeboard are distinct. This difference arises because the relative arrangement of the multi-rigid bodies and the direction of the constraints are decoupled into two degrees of freedom: the rotor and the wheel truck joints. This decoupling has two key implications. First, the efficiency of acceleration can be enhanced by alternating the rotation of the rotor and wheels to pump energy with the help of constraint forces. Second, in the steady state, energy savings can be achieved by applying compliant control to the rotor, allowing it to oscillate passively in tandem with the wheel oscillations. Similar to the roller racer, the magnitude of oscillation decreases as momentum increases. The two turning gaits, like those of the roller racer, are symmetrical to each other, and involve an additional cycle and a half added to the steady-state gaits to direct momentum for turning with minimal energy loss. In the demonstration, the turning motion occurs in two stages. The first stage serves as a preparatory motion, where the wheel and rotor are turned to one side. In the second stage, the actual large-angle turn is executed by turning both the wheel and rotor to the opposite side. Other gait transitions, including the start, transition, and end gaits, converge to the corresponding cyclic gaits using half cycles of appropriate amplitudes to maintain heading angles and match momentum.

The second row of Fig.~\ref{fig:simulation} presents the position trajectory and momentum history of the simulation results of a snakeboard executing the comprehensive motion plan starting from rest. During the acceleration phase, significantly larger oscillations are observed in both position and momentum compared to the steady-state phase, corresponding to the distinction between energy pumping and compliant control. The decoupled inertial manipulation and constraint direction control enable the snakeboard to achieve higher efficiency than the roller racer, but it also requires a longer acceleration time. Overall, these results demonstrate that the proposed method performs effectively in planning complex motions for the snakeboard.

\subsection{Intermediate Reynolds Number Swimmer}

\begin{figure}[!t]
\centering
\includegraphics[width=\linewidth]{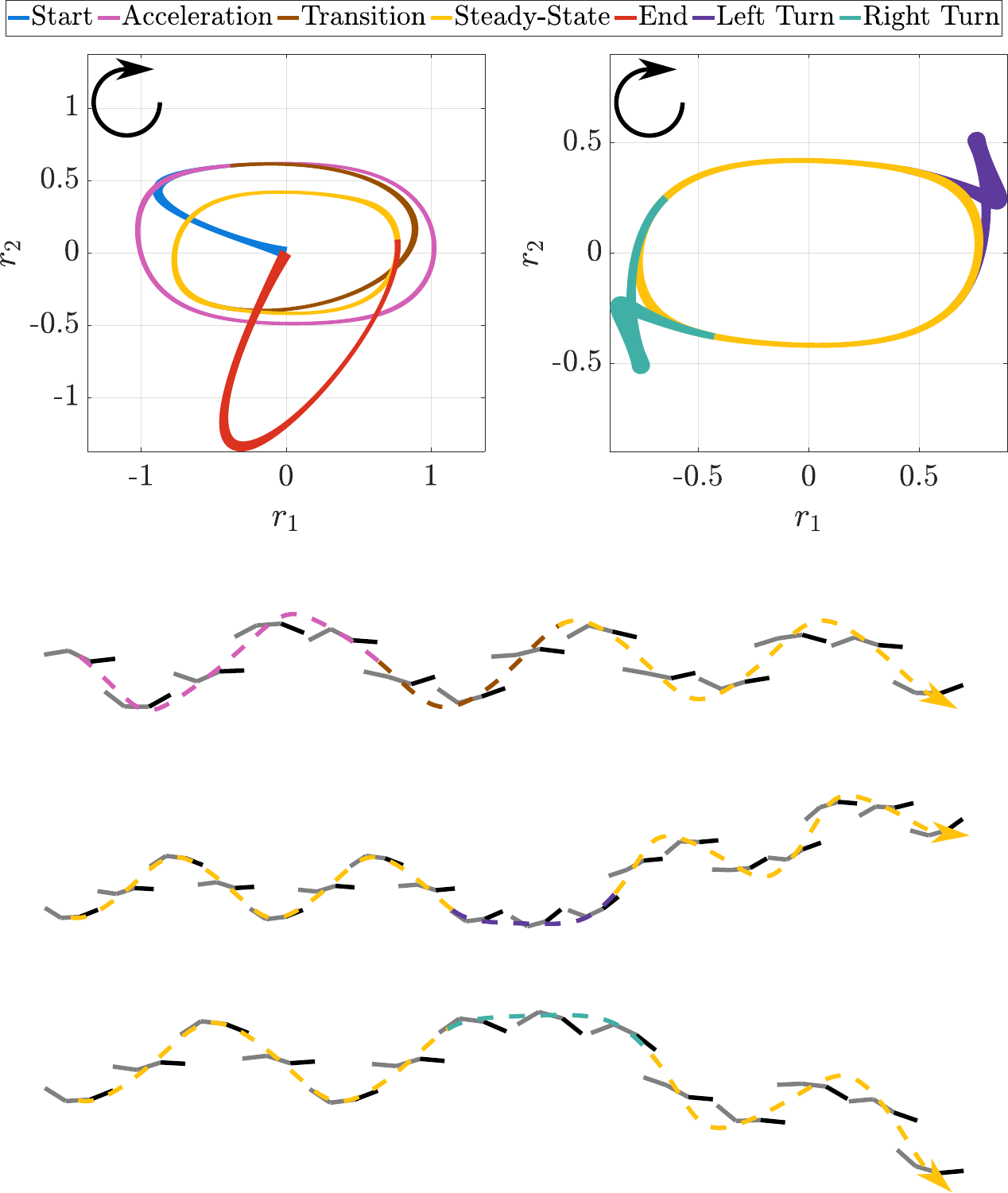}
\caption{Illustration of the optimal gaits and transitions for the intermediate Rynolds number swimmer (top) and their demonstration on the robot (middle: acceleration-transition-steady-state gait; bottom: steady-state-left-turn-steady-state and steady-state-right-turn-steady-state gaits). The thickness of the gait cycles indicates the pacing of the gait, with thick lines representing slower shape changes and thin lines representing faster changes. Black arcs with arrows denote the direction of the gait. In the demonstration, the dashed line represents the robot’s position trajectory, with its color indicating the corresponding gait being executed at each point. For clearer visualization, the swimmer's dimensions are scaled down from actual size.}
\label{fig:swimmer_gait}
\end{figure}

\begin{figure}[!t]
\centering
\includegraphics[width=\linewidth]{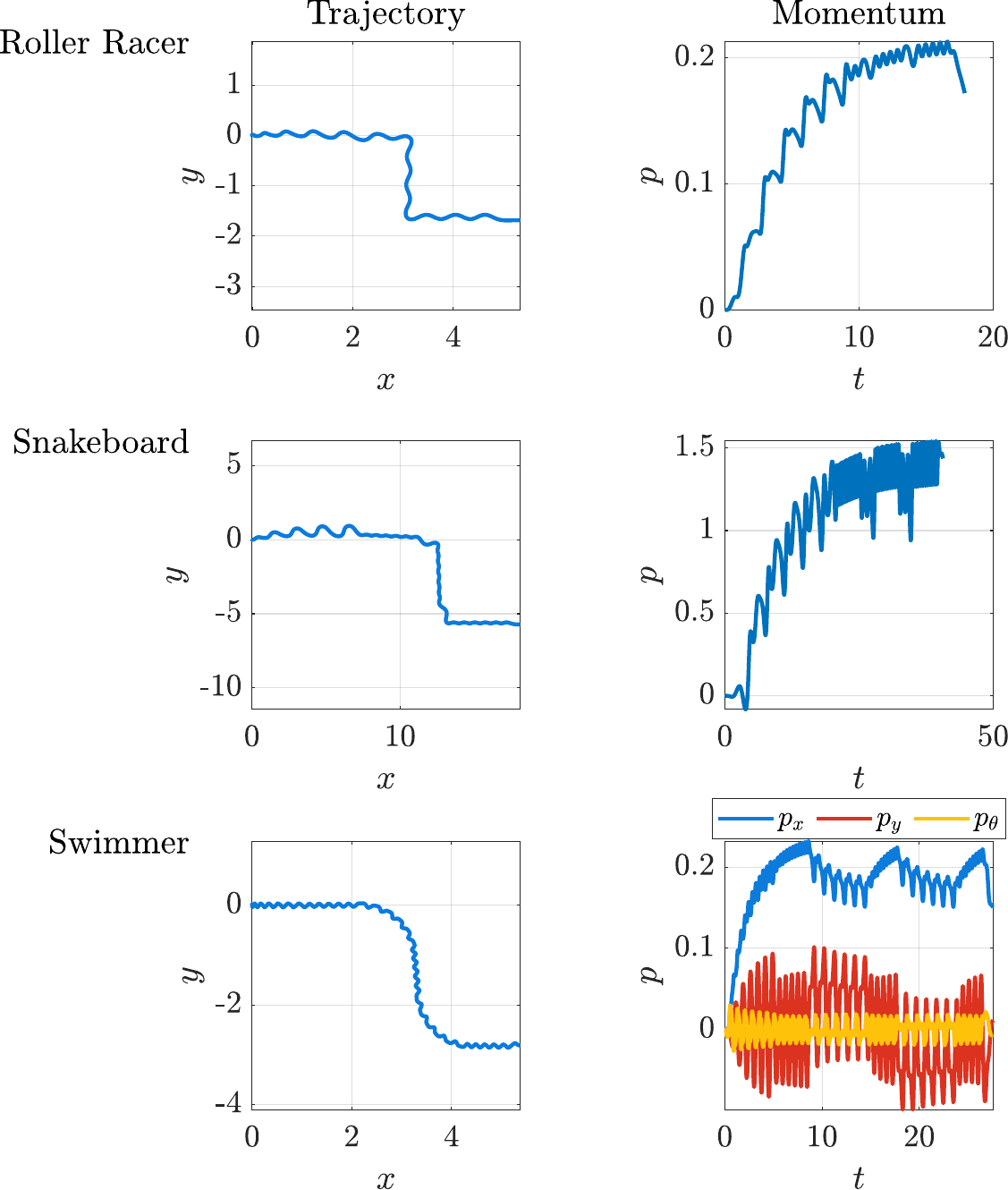}
\caption{Overall position trajectories and momentum-time curves from the execution of the integrated motion plan for three representative systems. The motion plan begins from rest, accelerates to the steady-state energy level, transitions to a steady state, performs a right turn, continues forward in the steady state, executes a left turn, continues forward in the steady state again, and finally returns to the nominal shape. For the turning gait, the swimmer turns 15 degrees per cycle, requiring six cycles to complete a quarter-angle turn.}
\label{fig:simulation}
\end{figure}

The intermediate Reynolds number swimmer is modeled as a three-link swimmer neutrally buoyant in the fluid with a total length of 0.6 meters. Each link is modeled as a cylindrical shape with a slenderness ratio of 1:5. Together with the actuator limits, this configuration results in swimming at a Reynolds number approximately between \( 2 \times 10^4 \) and \( 3 \times 10^5 \). Hydrodynamics are modeled using Morison's equation, with quadratic drag coefficients of 1.2 in the lateral direction and 0.008 in the axial direction \cite{orcina2025orcaflex}. Fluid-added mass is calculated based on the cylinder's cross-sectional geometry.

Fig.~\ref{fig:swimmer_gait} illustrates the optimal gaits and transitions for the swimmer in shape space and their demonstration on the robot. Similar to the snakeboard, the swimmer uses a larger-radius cycle for efficient acceleration and transitions to a streamlined steady-state gait to minimize drag. Notably, both the acceleration and steady-state gaits exhibit a larger amplitude for the tail joint (\( \base_1 \)) compared to the head joint (\( \base_2 \)). This difference likely arises from the need to maintain the heading angle by stabilizing the head while propelling forward with larger tail swings. Such behavior aligns with physical observations and analyses of fish propulsion, where wave amplitude increases toward the tail \cite{wu1961swimming}. As shown in the demonstration, turning is achieved by bending the body toward the desired direction, allowing drag forces to redirect momentum. Unlike the roller racer and snakeboard, which utilize constraint forces for turning, the swimmer actively drives its joints to generate turning forces through drag. This reliance on active actuation makes large-angle turning in a single cycle infeasible due to joint torque limits. Consequently, the desired turning angle is set to 15 degrees, requiring six cycles to complete a quarter turn.

The last row of Fig.~\ref{fig:simulation} shows the position trajectory and momentum history of the swimmer during the execution of the comprehensive motion plan starting from rest. Due to the six-cycle quarter-turn maneuver, the turning trajectory is less sharp compared to the roller racer and snakeboard. Additionally, the momentum and energy losses during turning are significantly higher than in the other systems. This difference stems from the swimmer’s reliance on large lateral drag forces for turning, which inherently dissipate substantial energy. Furthermore, as previously mentioned, although drag is quadratic with respect to the local body velocity, an efficient gait that alternates shapes correctly can minimize drag during swimming. This optimization results in an acceleration process that resembles the behavior of a linear mass-damper system, as shown in the momentum trajectory. Overall, the simulation results highlight the effectiveness of the proposed method in generating motion plans for the swimmer.

\section{Experiment Results}

We conducted experiments on the roller racer robot. The hardware consists of two pairs of passive skateboard wheels, each with a diameter of 0.8 m, attached to the front and rear carts. A Dynamixel XL430-W250-T actuator was used to control the joint between the two carts, with a bearing isolating the linear force. The battery, actuator, and bearing are all mounted on the rear cart, giving it greater inertia compared to the front cart. To evaluate pure feed‑forward performance, the motion plan was loaded onto a Raspberry Pi and executed in an open-loop manner, with the actuator solely tracking the desired shape. 
The viscous friction coefficient of the wheels was determined experimentally by pushing the robot and observing its deceleration until it came to a stop due to friction. The robot's position trajectory was tracked using Aruco markers attached to the robot and the environment.  

Fig.~\ref{fig:intro} illustrates the robot's movement and behavior when executing the motion plan from rest. The composite plots show that the robot successfully reproduces the planned behaviors, including starting from rest, accelerating, transitioning between gaits, maintaining steady-state motion, turning, and returning to the zero shape to glide. These results align closely with the simulation outcomes. Given that the gaits were executed in an open-loop manner, this performance demonstrates the validity and robustness of our approach.  

\begin{figure}[!t]
\centering
\includegraphics[width=\linewidth]{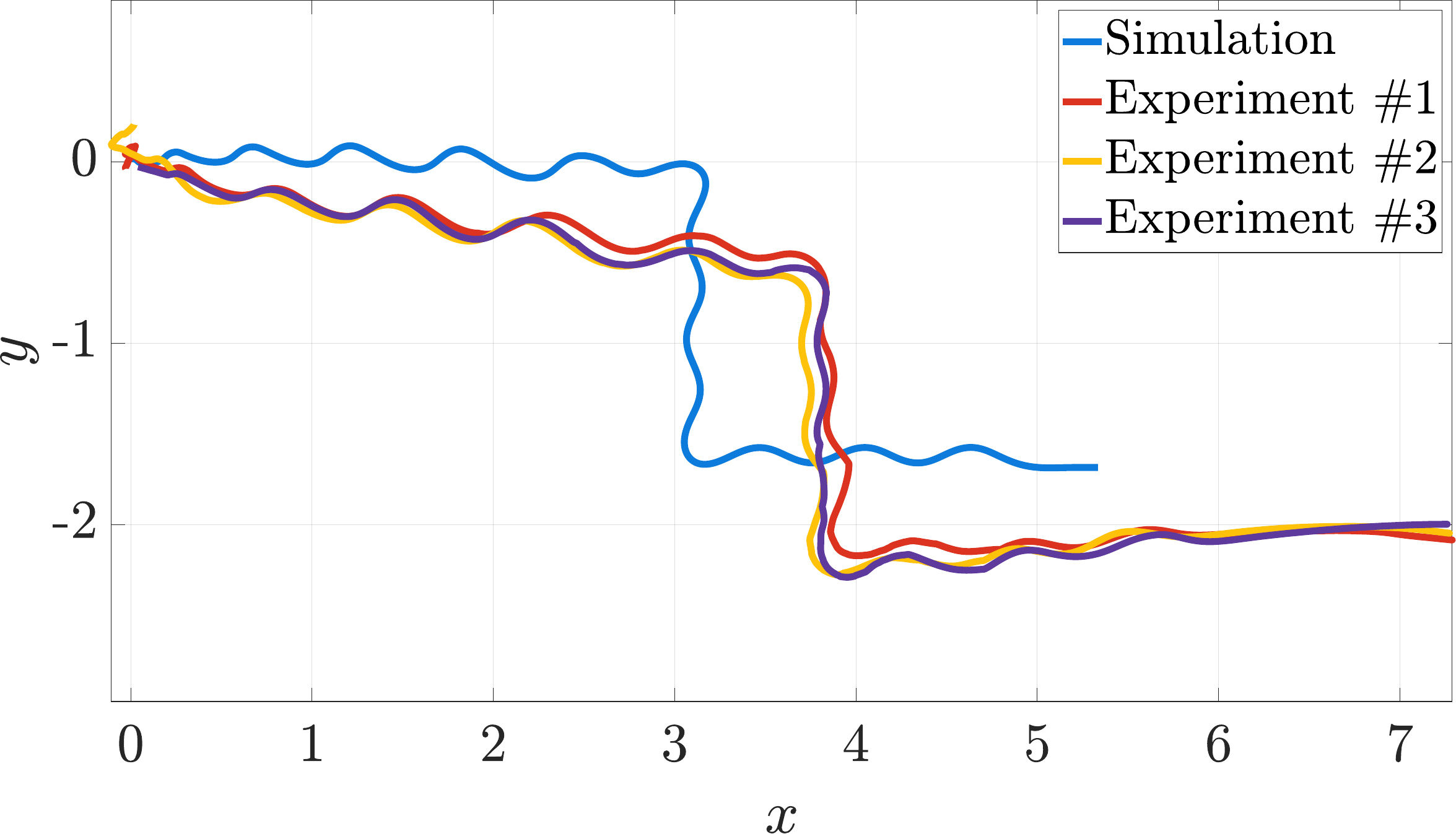}
\caption{Comparison of experimental and simulated position trajectories for the roller racer robot. Three open-loop experiments were conducted to execute the comprehensive motion plan derived using the proposed method. The experimental trajectories were obtained using Aruco markers on the roller racer and within the environment. A low-pass filter was applied to the experimental results to reduce noise.}
\label{fig:experiment}
\end{figure}

Fig.~\ref{fig:experiment} presents the position trajectories of three experimental trials for quantitative analysis. The results show that the robot effectively reproduces the overall motion and gait behaviors, despite exhibiting approximately 10 degrees of orientation drift at the beginning of the acceleration and 2 meters of additional displacement compared to the simulation. The primary sources of error are likely due to model mismatches, including friction not being purely viscous but potentially involving Coulomb friction, imperfect shape tracking, inaccuracies in the robot's inertia computation, and possible wheel sliding at high speeds. Because the experiments were conducted in an open-loop setup, these errors accumulated and scaled up over time. Despite these errors, the experimental results demonstrate that the feed-forward motion plan and gaits computed using the proposed approach can be successfully executed on the robot hardware, achieving the desired movements with promising performance---even in open-loop control. Furthermore, the consistency across the three experimental trials highlights the robustness of our approach.

\section{Limitations}

While our work demonstrates effectiveness and robustness in both simulations and experiments across three example systems, several limitations remain and provide opportunities for future work. First, the framework currently generates only feed‑forward gaits and motion plans; it does not incorporate closed-loop control. In practice, mitigating real-world errors requires feedback. For example, the feedback controller based on averaging theory in \cite{vela2002second, vela2003averaging} can complement our feedforward gait, as both are derived from Lie group theory. In addition, implementing feedback would involve online state estimation, for instance via the invariant Kalman filter \cite{barrau2018invariant}, using IMUs and wheel encoders.

Second, as evidenced by the experimental results, the current model based on viscous friction lacks the fidelity to fully capture real robot behavior. Addressing this limitation requires adopting higher-fidelity models for both wheeled robots and swimmers. Potential approaches include improved friction models \cite{salman2016physical}, boundary element methods for coupled fluid-added mass computations \cite{kanso2005locomotion}, or system identification techniques \cite{justus2024geometry}. 

In addition, the limited turning angle achievable within a single cycle for the swimmer highlights a constraint in the current gait design. Potential solutions include developing turning gaits independent of the steady-state gait to improve efficiency. Alternatively, addressing the fundamental limitation---that swimmers cannot leverage constraint forces for turning, unlike the roller racer or snakeboard---could involve designing mechanisms such as fins or rudders to utilize constraint forces for energy-efficient turning \cite{morgansen2007geometric}. Additionally, transitioning from articulated designs to continuum swimmers with soft actuators could enable more efficient bending and propulsion~\cite{yang2025geometric}. 

Moreover, the current approach assumes a single gait is used throughout the acceleration process. While this simplifies implementation, it sacrifices some efficiency. A potential direction for future work is to dynamically adjust the acceleration gait during the process based on momentum changes. This could be achieved by tracking the solution of the time-varying acceleration gait optimization problem using an online algorithm, such as those described in \cite{simonetto2017predictioncorrection, choi2022optimal}. 

Finally, the Lie group integrator discretization used in this paper employs a method analogous to the forward Euler method on the tangent space, which may be less accurate compared to higher-order methods, such as the Cayley mapping proposed in \cite{marsden2001discrete, kobilarov2011discrete}. Future work will focus on extending the discretization of Lie group integrators to higher-order methods to enhance accuracy and performance.

\section{Conclusion}

In this paper, we propose a gait optimization and motion planning framework for kinodynamic systems. The framework begins with a general kinodynamic model that captures both kinematic and dynamic effects using Lagrangian reduction and differential geometry. Building on this model, we introduce a variational gait optimization approach based on Lie group integrators and group symmetries. Finally, we construct motion plans by combining diverse gaits and transitions to achieve various motions, including acceleration, maximizing steady-state speed, turning, and transitions. The proposed framework was evaluated on wheeled robots with nonholonomic constraints and swimming robots with anisotropic fluid-added inertia and hydrodynamic drag, demonstrating consistent performance and robustness across both simulations and hardware experiments. These results underscore the validity and effectiveness of our approach.

\bibliographystyle{plainnat}
\bibliography{references}

\end{document}